\documentclass[11pt,final]{article}
\usepackage{amsfonts}
\usepackage{amsmath}
\usepackage{amssymb}
\usepackage{amsthm}
\usepackage[margin=1.0in]{geometry}
\usepackage{graphicx}

\def\colb{}

\def\R{{\mathbb{R}}}

\def\dt{{\Delta t}}

\def\Xh{X}
\def\rt{{\tilde{r}}}

\def\calD{{\cal D}}
\def\calT{{\cal T}}

\def\bX{{\mathbf{X}}}
\def\bR{{\mathbf{R}}}

\theoremstyle{thmstyleone}%
\theoremstyle{thmstyletwo}%

\theoremstyle{thmstylethree}%

\begin{document}

\title{Predicting Shallow Water Dynamics using Echo-State Networks with Transfer Learning}

\date{\today}

\author{
Xiaoqian Chen\thanks{\texttt{University of Houston, Department of Mathematics, chenxq0709@gmail.com}}~,
~~Balasubramanya T. Nadiga\thanks{\texttt{Los Alamos National Laboratory, balu@lanl.gov}}~,
~~Ilya Timofeyev\thanks{\texttt{University of Houston, Department of Mathematics, ilya@math.uh.edu}}
}

\maketitle

\abstract{In this paper we demonstrate that reservoir computing can be used to learn the dynamics of the shallow-water equations.
  In particular, while most previous  applications of reservoir computing have required training on a particular trajectory to further predict the evolution along that trajectory alone, we show the capability of reservoir computing to predict trajectories of the shallow-water equations with initial conditions not seen in the training process.
  However, in this setting, we find that the performance of the network deteriorates for initial conditions with ambient conditions (such as total water height and average velocity) that are different from those in the training dataset.
  To circumvent this deficiency, we introduce a transfer learning approach wherein a small additional training step with the relevant ambient conditions is used to improve the predictions.}

\medskip



\section{Introduction}
\label{sec:intro}

Recent advances in computational capabilities and availability of extensive observational or numerical datasets rekindled interest in developing various machine learning (ML) methods for predicting and understanding dynamics of complex systems. Performing direct numerical simulations (DNS) of complex partial differential equations (PDEs) representing multiscale processes can be computationally prohibitive in many practical applications, especially if ensemble simulations are required to predict evolution of averaged quantities or assess uncertainty. The idea of replacing computationally expensive direct numerical simulations with simpler models has been studied extensively in the literature. Semi-analytical examples include various homogenization techniques, developing large-eddy simulation models, 
multigrid numerical methods, etc.
Machine Learning offers a purely empirical approach for estimation of computationally 
simpler (reduced, effective) models which are capable of reproducing behavior of more complex systems.

Many novel ML methods have been introduced and developed recently with the emphasis on predicting solutions of chaotic systems or learning the underlying dynamic equations. Some of the most popular approaches for time-series analysis and prediction utilize sequential models such as artificial neural networks (ANNs) that include feedforward neural networks (FNN) and recurrent neural networks (RNN).
RNN architectures are characterized by "loop" connections  between individual neurons, and it is well understood that it can be rather difficult to train general RNNs because of their complex connectivity. 
Therefore, a novel Reservoir Computing (RC) approach was 
introduced recently to overcome this shortcoming of general RNNs 
\cite{jaeger2001echo, maass2002real}. In this approach, the reservoir consists 
of a fixed RNN with randomly selected connections and weights, and only the output matrix is trained to match the data. Thus, training is very straightforward in this context since it is equivalent to linear convex optimization problem with an explicit solution.
It has been demonstrated that RC performs very well in the context of predicting chaotic dynamics (see, for example, \cite{jaeger2004harnessing, pathak2018model, chattopadhyay2020data, vlachas2020backpropagation, bollt2021, nadiga2021reservoir})
and this approach has also been successfully applied in the context of climate and weather prediction \cite{nadiga2021reservoir, arcomano2020machine}, turbulent convection \cite{pasch20}, etc.
{\colb Comparison of Reservoir Computing with other ML methods for predicting spatiotemporal 
chaotic dynamics is presented in 
\cite{vlachas2020backpropagation, chattopadhyay2020data}.
}

In this paper we apply the RC methodology to learn the dynamics of the Shallow Water Equations (SWE). To this end, we extend previous results for the reservoir computing beyond predicting just one trajectory. In particular, we develop a methodology to predict SWE trajectories with initial conditions which are not in a training dataset and even solutions with different ambient physical properties (e.g. the total mass) than that of the training data.

Although the shallow water equations considered here are not chaotic, the solutions of the system exhibits complex transient behavior. 
We demonstrate that RC is capable of learning the SWE dynamics and 
accurately reproducing transient behavior of the SWE model.
However, large scale quantities such as the averaged water height and velocity play an important role, and RC performs much better for solutions with initial conditions which are consistent 
(in terms of these quantities) with the training dataset. 
Performance of the RC method deteriorates quickly for initial conditions with large-scale quantities which are not represented in the training regime. Thus, we introduce a transfer learning approach to re-train the RC output matrix on an extremely small dataset generated by a single short simulation of the SWE. We demonstrate that this approach drastically improves the performance of the RC method.

\section{Methods}
\label{sec:methods}

\begin{figure}[h!]
    \centering
    \includegraphics[width=0.9\textwidth]{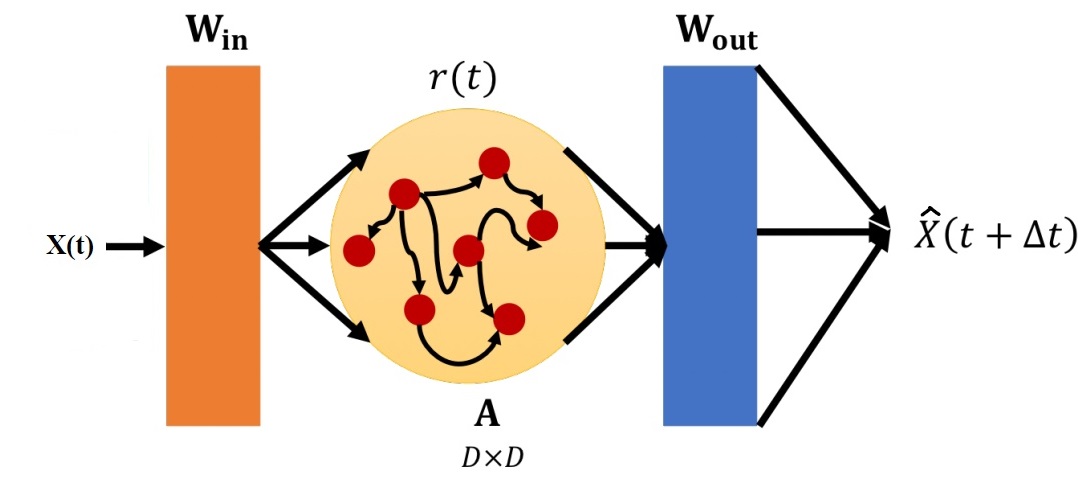}
    \caption{A schematic of Reservoir Computing architecture .
    Input and adjacency matrices $W_{in}$ and $A$ are fixed random matrices. During training,  only the output matrix $W_{out}$ is computed via an optimization procedure using training data. In an autonomous setting, during prediction, the output $\hat{X}(t+\dt)$ is computed and then used as input for the next time-step.}
    \label{fig1}
\end{figure}

In this section we give a quick overview of Reservoir Computing and discuss a particular implementation of this method using Echo-State Networks (ESN). 
Figure \ref{fig1} schematically describes the idea behind the reservoir computing. Here we concentrate on the prediction setting where given a dynamic variable 
$X(t) \in \R^N$ we use ESN to predict the future value of the trajectory $\Xh(t+\dt)  \in \R^N$. The RC method consists of the reservoir $r(t) \in \R^D$
described by the reservoir matrix $A \in \mathbb{R}^{D \times D}$ with ``echo state property'' \cite{yildiz2012re}, input connectivity matrix $W_{in} \in \R^{D \times N}$, and output matrix $W_{out} \in \mathbb{R}^{N \times D}$. 
The update of the internal reservoir state $r(t)$ is given by
\begin{equation}
\label{eq:res_state}
    r(t+\Delta t) = f \left( A \, r(t) + W_{in} X(t) \right),
\end{equation}
where
$f: \R^D \to \R^D$ is the element-wise application of a non-linear
activation function ($f \equiv \tanh$ in our case).
It was demonstrated in \cite{pathak2018model, chattopadhyay2020data} that the nonlinear transformation can potentially increase the expression power of the reservoir.
We compared several nonlinear transformations in 
\cite{xiaothesis} and demonstrated that the transformation 
\begin{align}
\label{rt}
    \rt_j(t) &=  \begin{cases}
                    r_{j}^2(t) & \text{if $j$ is odd},\\
                    r_{j}(t) & \text{if $j$ is even}
                \end{cases}
\end{align} 
slightly increases the performance of the RC method.
Thus, the prediction equation becomes 
\begin{equation}
\label{eq:predict}
\Xh(t + \Delta t) = W_{out} \rt(t+\Delta t),
\end{equation}
where $\rt(t)$ is the modified state of the reservoir defined in
\eqref{rt}.
Equations \eqref{eq:res_state} and \eqref{eq:predict} can be utilized sequentially
to generate future predictions $\Xh(t + 2\Delta t)$, $\Xh(t + 3\Delta t)$, etc. Thus, RC equations \eqref{eq:res_state} and \eqref{eq:predict} 
are used here as, essentially, a time-stepping algorithm.
Often, prediction starts with initial condition $X(0)$ which corresponds to the initial reservoir state $r(0)=0$. Various improvements have been suggested (see \cite{xiaothesis} for a detailed comparison), such as reservoir warm-up which uses a short "true" trajectory to update the internal reservoir state using equation \eqref{eq:res_state}.

{\colb
Matrices $W_{in}$ and $A$ are generated at random beforehand and are kept constant during training and subsequent predictions.
Here we use $W_{in}[iq:(i+1)q,i] \sim Uniform[-\beta_1, \beta_1]$ for $i=0,N-1$, $q=D/N$. Thus, each input $X_i(t)$ drives only $q$ neurons in the reservoir.
Results are weakly sensitive to the value of $\beta_1$. In particular, $\beta_1$ should be in the proper range
so that non-linearity of $f$ in \eqref{eq:res_state} becomes essential. We use $\beta_1=0.1$ and also demonstrate prediction results for
$\beta_1 \in [0.01,\ldots,0.2]$ for a particular dataset. }
Connections within the reservoir are represented using the adjacency matrix $A \in \mathbb{R}^{D \times D}$ with ``echo state property'' \cite{yildiz2012re}. The echo state property is a condition of asymptotic state convergence of the reservoir network, under the influence of driving input. Intuitively, the echo state property states that the memory of initial conditions should decay 
asymptotically with time. A common practical approach is to select the spectral radius of $A$ 
smaller than unity to ensure the echo state property \cite{jaeger2001echo, yildiz2012re, buehner2006, jaeger2007optimization, Jaeger2013}.
It has been suggested \cite{esp1} that the condition on the spectral radius may not be sufficient to ensure the echo-state property. Moreover, it has been also suggested that the echo-state property can be considered as input-dependent and in such cases the echo-state property might also
hold for the adjacency matrix $A$ with the spectral radius larger than unity \cite{esp3}.
Nevertheless, it has been demonstrated that 
the condition on the spectral radius of $A$ 
remains a valid indicator of the echo-state property for a large class 
of networks \cite{esp2} and, thus, we follow this approach here.
$A$ is usually a sparse matrix. Thus, we first generate a sparse matrix 
$W_0$ (usually less than 10\% of connections) with entries 
drawn from a uniform distribution with zero mean. 
Then, $A$ is given by
\begin{equation}
    A = \beta_2 \frac{W_0}{\vert \Lambda_{max} \vert},  
    \label{eq:A}
\end{equation}
where $\vert \Lambda_{max} \vert$ is the largest eigenvalue (in absolute value) of $W_0$ and $\beta_2 \le 1$ is a scaling parameter.
Thus, the spectral radius of A is $\rho(A)  = \beta_2$.
The optimal value of $\rho(A)$ should be set depending on the desired amount of memory and non-linearity for a particular task. As a rule of thumb discussed in \cite{jaeger2007echo, jaeger2001echo, yildiz2012re}, $\rho(A)$ should be close to 1 for tasks that require long memory, and $\rho(A)\ll 1$ for tasks where a long memory might be harmful. 
Larger $\rho(A)$ also have the effect of driving signals $X(t)$ into more nonlinear regions of $\tanh$, similarly to larger $W_{in}$. Thus, scaling of both $W_{in}$ and $A$ have a similar effect on the degree of non-linearity of the ESN, while their difference determines the length of the memory effect.
In \cite{xiaothesis} we demonstrated that prediction utility is affected weakly by the spectral radius in the range $\beta_2 \in [0.01, 
\ldots, 0.9]$ with a slightly better performance for $\beta_2=0.1$.
Thus, we use $\beta_2=0.1$ in this work.

Only weights of the output layer $W_{out} \in \R^{N \times D}$ 
are updated during training. These weights are computed as a linear regression with $L^2$ regularization
\begin{equation}
    W_{out} = \underset{W}{\mathrm{arg min}} \; || W \bR - \bX ||_2^2 + \lambda || W ||_2^2 ,
    \label{eq:wout}
\end{equation}
where $\bR = \{\rt(j\dt), j=0,\ldots,T\}$, $\bX = \{X(j\dt), j=0,\ldots,T\}$,
and
$\lambda$ is the $L^2$ regularization (ridge regression) parameter
also known as Tikhonov regularization \cite{golub1999tikhonov}.
It can be shown that 
$W_{out}$ can be computed explicitly as
\begin{equation}
    W_{out}' = (\bR \bR' + \lambda I)^{-1} \bR \bX'.
\end{equation}
Here, $( \cdot)'$ is the transpose and $(\cdot)^{-1}$ is the inverse operation.
This implies that training of the ESN can be done quite fast
compared to other neural networks.
In this work we construct the matrix $\bX$ by concatenating
several trajectories of the SWE with different initial conditions. 
Matrix $\bR$ is then constructed by using the equation \eqref{eq:res_state}
with the "true" trajectory $\bX$.
However, different initial conditions are generated with 
identical ambient physical constraints for the averaged water height and averaged velocity. We comment further on the training dataset in the next section.

The initial state of the reservoir is $r(0)=0$.
In \cite{xiaothesis}, we demonstrated that reservoir warm-up can improve ESN prediction, especially over short initial times. 
To this end, one has to perform a short DNS of the shallow-water equations to generate a "true" trajectory and update the reservoir dynamics for a few initial time-steps using \eqref{eq:res_state}. 
However, here we mimic a more realistic situation when the "true" trajectory is not available.

\textbf{Transfer Learning.}
{\colb The general idea of transfer learning 
is usually formulated using the source and target domains and tasks. Assuming that there is a source domain $\calD_s$ (this is data $\text{TEST}_0$ in our case) a task $\calT_s$ (predicting trajectories consistent with $\text{TEST}_0$) 
can be learned using supervised learning (training the ESN and computing $W_{out}$). However, it is often desirable to learn a new task, 
$\calT_t$ (predicting trajectories not consistent with ambient physical constraints in $\text{TEST}_0$), in a target domain $\calD_t$ (trajectories in the corresponding $\text{TEST}_j$ dataset) with $\calD_t \ne \calD_s$.
Transfer learning aims to improve the target task by using some knowledge from the source domain. Without transfer learning, ESN trained in the $\calD_s$
would be discarded and an ESN in the $\calD_t$ needs to be trained from scratch.
Obviously, this requires a large amount of data generated in the $\calD_t$ domain.
Transfer learning avoids this by using information in the $\calD_s$ 
($W_{out}$ computed using $\text{TEST}_0$ training data) and using
only a small amount of data in the $\calD_t$ domain to modify previously computed 
$W_{out}$.

In terms of predicting time-series data, transfer learning can help to utilize an NN model outside of the parameter 
regime where it was originally trained. }
This can be a particularly challenging task in the context of dynamical systems. For instance, even if a dynamical system is ergodic, the shape of the attractor might depend on conserved quantities, value of the viscosity coefficient, etc. Therefore, performance of NN models outside of their training regime can be quite poor. Recently, it has been recognized that transfer learning can remedy this situation
(see e.g. \cite{inubushi2020transfer}). To this end, an additional small dataset has to be generated with new parameters, and "old" NN weights can be modified using new data. Thus, transfer learning makes use of the previously learned model while adjusting it quickly for a new parameter regime. Moreover, for the transfer learning to be efficient, 
the new dataset should be much smaller than the original training data.

If we assume that an additional training data, $\bX^*$, is available,
then the modified weights $\tilde{W}_{out} = W_{out} + \delta W$ 
are determined by the optimization problem
\begin{equation}
\label{eq:deltaw1}
    \delta W = \underset{\delta W}{\mathrm{arg min}} \; || (W_{out} +\delta W) \bR^* - \bX^* ||_2^2 + \alpha || \delta W||_2^2
\end{equation}
and
the corresponding analytical solution for the correction is
\begin{equation}
\label{eq:deltaw}
   \delta W'  =(\bR^* (\bR^*)'+ \alpha I)^{-1} (\bR^* (\bX^*)'- \bR^* (\bR^*)' W_{out}'),
\end{equation}
where $\alpha$ is the transfer learning rate.
When the transfer rate is zero, $\alpha= 0$, the above formula reduces to the conventional training method with $\lambda=0$, which is just supervised learning by using the target training data $\bX^*$ only. There is no knowledge transfer from the source domain for $\alpha= 0$. On the other hand, in the limit of large transfer rate, $\alpha \to \infty$, we obtain $\delta W \to 0$, since there is a strong penalty on $\delta W$. Thus, for $\alpha \to \infty$ the above formula implies reusing weights $W_{out}$ without any correction, i.e., $\delta W=0$. 
Thus, there is no knowledge gain from the target domain for $\alpha \gg 1$. 
Therefore, the goal of transfer learning approach is to select an appropriate value of $\alpha$ which balances information from the previously learned model and new data.
{\colb If the dimension of new data 
$\bX^* = \{X^*(j\dt), j=0,\ldots,T^*\}$
is small, i.e. $T^* \ll D$, then the matrix 
$\bR^* = \{\rt^*(j\dt), j=0,\ldots,T^*\}$ has dimensions 
$\bR^* \in \R^{D \times T^*}$ and $\text{rank}(\bR^*) \le T^*$. 
Therefore,  $\bR^* (\bR^*)' \in \R^{D \times D}$ is a low-rank matrix and 
$\alpha$ acts as a regularization parameter.
}

\section{Shallow-Water Equations}
\label{sec:swe}

The Shallow Water Equations (SWE) are a widely-used model to describe fluid flow in rivers, channels, estuaries or coastal areas. The main assumption of the shallow water model is that the horizontal length scale is much greater than the depth scale. 
In one dimension, the shallow water equations take the following form

\begin{equation} \label{swe}
\begin{split}
 & \partial _t h + \partial _x (hu) =0 , \\
 & \partial _t (hu) + \partial _x \left(hu^2+ \frac{1}{2}gh^2 \right)+ gh \partial _x z - \nu \partial_{xx}(hu)=0 ,
\end{split}
\end{equation}
where
$h(x, t)$ is the water height, 
$u(x, t)$ is the fluid velocity,
$g$ is the gravitational constant,
$\nu > 0$ is viscosity constant,
$z(x)$ is the bottom topography, which does not depend on time.
We use periodic boundary conditions $h(x,t) = h(x+L,t)$, $u(x,t) = u(x+L,t)$.
We add viscosity in equations \eqref{swe} to control small-scale
numerical oscillations arising in long-term simulations.
{\colb
Here we consider the non-dimensional form of the equations with $g=32$. In the dimensional form
the gravitational acceleration is $g=9.81 m/s^2$. To obtain the non-dimensional version, one has to rescale the equations and constant $g$ by the suitable units of space and time. Obviously, there are many combinations of space and time rescaling factors which result in the same non-dimensional version of the equations. Just to provide some intuition, $g=32$ corresponds approximately to, for example, dimensional units $[L] = 1\text{km}$ and $[t] = 1\text{min}$.
}

{\colb On the other hand, if we choose the reference time as the
  advection time for the flow to traverse the domain $[t] = L/U$, the
  viscous SWE has two relevant 
  non-dimensional numbers the Froude number and the Reynolds number
  and the equations may be written in the following non-dimensional
  form:
\begin{equation} \label{swe-nd}
\begin{split}
 & \partial _t h + \partial _x (hu) =0 , \\
 & \partial _t (hu) + \partial _x \left(hu^2+ \frac{h^2}{F\!r^2} \right)+ \frac{1}{F\!r^2}h \partial _x z - \frac{1}{Re} \partial_{xx}(hu)=0 ,
\end{split}
\end{equation}
where the Froude number $F\!r$ is given by  $F\!r = \frac{U}{\sqrt{gh_0}}$
and the Reynolds number $Re$ is given by $Re = \frac{UL}{\nu}$. For
the reference conditions indicated in the table below, the Froude
number corresponds to $F\!r$ = 0.221 and the Reynolds number
corresponds to $10^5$. That is, we consider a flow that is subcritical
(flow speed less than the long gravity wave propagation speed) and
that is mildly frictional.}

Trajectory prediction using ESN has been primarily applied to chaotic systems with a unique underlying attractor (see e.g. \cite{pathak2018model, chattopadhyay2020data, pyle2021domain, inubushi2019transferring, nadiga2021reservoir}).
Lyapunov exponents are typically used as a measure of instability
of small perturbations. However, the attractor of the equations
in \eqref{swe} is rather simple. 
{\colb
It appears numerically that in the presence 
of viscosity all trajectories for long times converge to a stationary state which corresponds to the
"flow over the bump" profile. }
Therefore, we use finite-time Lyapunov exponents to quantify the behavior of near-by trajectories for finite times. In particular, the largest finite-time Lyapunov exponent  
is positive for short times, but becomes close to zero (or even small and negative) for larger times.
Such behavior is likely related to the phenomenon of transient growth of perturbations in an asymptotically stable system.
The latter, in turn, can occur when the eigenvectors of the system are non-normal, a situation that commonly arises when the underlying state of the system is non-uniform.
{\colb
For short times SWE dynamics exhibits interesting transient behavior and we concentrate on predicting trajectories on the interval $t\in[0,20]$. We can extend predictions of the ESN further in time, but that becomes rather straightforward since solutions converge to a steady state profile.} 

We use Godunov-type numerical scheme for the SWE developed in \cite{acu15}.
The scheme is well-balanced, entropy-preserving in the absence of viscosity, and preserves the ``lake at rest'' equilibrium solution.
We utilize a setup considered in many publications including \cite{goutal1997proceedings}. In particular, we consider the 
"bump" topography in the middle of the domain $[0, L]$
\begin{align}
    z(x) &=  \begin{cases}
                   H \left( 1- \left(\frac{x- 0.5L}{ 0.5 W} \right)^2\right) & \text{if $|x-0.5 L|\le 0.5 W$}\\
                    0 & \text{otherwise},
                \end{cases}
\end{align}
where $H$ and $W$ are the height and width of the topography, respectively.

In this paper we consider a reference setup where initial conditions 
are generated as perturbations of the flat initial water height
and constant initial velocity
\begin{equation}
\label{icswe}
    h(x, t=0)+z(x) = h_0, \qquad u(x, t=0)= u_0.
\end{equation}
Parameters for direct numerical simulations of the SWE are presented in the table below, where $\Delta x$ and $\delta t$ are space and time discretizations, respectively. 
{\colb  Time step $\Delta t$ is used to generate data 
for the training of the ESN and generating prediction trajectories. }
\begin{center}
\begin{tabular}{|p{0.6cm}|p{0.6cm}|p{0.7cm}|p{0.6cm}|p{0.6cm}|p{0.6cm}|p{0.6cm}|p{0.6cm}|p{0.9cm}|p{1.4cm}|}
\hline
 L & g & $\nu$ & $h_0$ & $u_0$ & $H$ & $W$  & $\Delta x$ & $\delta t$ & ESN $\Delta t$\\
\hline
40 & 32 &  0.001 & 4.0 & 2.5   & 0.48 & 8.0 &  0.1   &0.0005 & 0.1\\
\hline
\end{tabular}
\end{center}
We consider the reservoir size $D=4800$ and both the water height and momentum are discretized with $n=400$ points. Since we consider predictions for both, water height and momentum, $N=800$ in the definition of the input and output matrices $W_{in}$ and $W_{out}$
discussed in section \ref{sec:methods}. 
{\colb  The CFL condition for the parameters and regimes considered here is
$\delta t < O(10^{-3})$ (exact constant depends on the velocity and water height level).
Therefore, ESN predictions are generated with the time-step $\Delta t$ which is approximately 
100 times larger than the largest possible time-step in numerical simulations of the SWE.}

Equations in \eqref{swe} conserve the mean water height, $h_0$, which
corresponds to the conservation of mass. Although momentum is no
longer conserved because of the presence of bottom topography (form
drag or pressure drag), momentum changes very slowly in the
simulations. In particular, for simulation times presented here
changes in the momentum are less than 0.2\%.  

{\bf Training Dataset.} 
Since our goal is to learn the dynamics of the shallow-water model, we do not concentrate on a single trajectory. Instead, we compose the training dataset using several trajectories with different initial conditions. In particular, we simulate the SWE dynamics \eqref{swe} numerically to generate $M=20$ trajectories on $t\in[0,20]$ with random initial conditions 
\begin{equation}
\label{IC_train}
\begin{split}
&h(x, 0) +z(x)  = h_0 + a \, h_0 \, \sin(2 k \pi x /L + \omega_1) ,\\
&u(x, 0)=u_0 + d \, u_0 \, \sin(2 p \pi x /L + \omega_2) ,
\end{split}
\end{equation}
 where parameters $h_0=4$ and $u_0=2.5$ are the mean water height and velocity.
 Parameters $h_0$ and $u_0$ are identical for all trajectories in the training data and different initial conditions are obtained by sampling other parameters from various uniform distributions. 
 In particular, 
 both $a, \, d  \sim Uniform[0, 0.05]$. Frequency for the perturbation are discrete uniform random variables with $k, p \sim Uniform(1,2,3,4)$ and $\omega_1, \omega_2 \sim Uniform[0, 2\pi]$. The sampling time-step for collecting the data is $\Delta t =0.1$. We concatenate $M$ trajectories to obtain the training data $\bX$ and then use the internal reservoir dynamics \eqref{eq:res_state} to compute $\bR$. Thus, each time-instance of $\bX$ consists of a vector of size 
 $2n = 2L/\Delta x$ (snapshots of $h(j\Delta x,t)$ and $uh(j\Delta x,t)$, $j=1,\ldots,n$), 
 where $n$ is the number of points in the spatial discretization. The concatenation of different trajectories leads to discontinuities in the training data and therefore in the reservoir dynamics.
 However, there are only $M=20$ discontinuous points in the training data vs approximately 
 $4,000$ points of continuous sampling. Therefore, $20$ discontinuous points degrade the training of the ESN slightly, but it does not have a strong effect on the prediction power of the ESN.
 In particular, there are no jumps in the predicted trajectories at
 time of the discontinuity in the training data (e.g. at $t=20$).
%
 %

{\bf Testing Dataset.} 
Testing data consists of two sets of trajectories with identical sets of initial conditions. One set of trajectories are the "true" trajectories obtained by numerically simulating the SWE dynamics in \eqref{swe}. The second set of trajectories consists of "predicted" trajectories 
obtained using the ESN. The "predicted" trajectories are generated with the
time-step $\Delta t = 0.1$.

Similar to the training data, we generate "true" and "predicted" sets of $J=20$ 
trajectories on $t\in[0,20]$ with random initial conditions
\begin{equation}
\label{IC_test}
\begin{split}
 &h(x, 0) +z(x)  = h_0 + s_h + a \, h_0 \, \sin(2 k \pi x /L + \omega_1) ,\\
 &u(x, 0)= u_0 + s_u + d \, u_0 \, \sin(2 p \pi x /L + \omega_2) ,
\end{split}
\end{equation}
where we introduced "shift" parameters $s_h$ and $s_u$ which affect the initial mean water height and 
velocity. Thus, these "shift" parameters induce changes in the ambient large-scale quantities (e.g. initial total mass and momentum).
Similar to the training data $a, \, d  \sim Uniform[0, 0.05]$, 
$k,\, p \sim Uniform(1,2,3,4)$, 
and $\omega_1, \omega_2 \sim Uniform[0, 2\pi]$.
In \cite{xiaothesis} (section 7.1) we demonstrated that perturbations of higher wavenumbers decay faster than perturbations of low wavenumbers. {\colb 
Therefore, prediction errors are approximately twice smaller and decay faster for trajectories with initial conditions with perturbations in higher wavenumbers. 
Thus, we only consider trajectories with initial perturbations with wavenumbers $k, p = \{1,2,3,4\}$ for testing, which is a more challenging set of initial conditions for predicting the dynamics of the SWE.
We comment about initial conditions with higher frequencies in Section \ref{sec:results}.
}

In this work, we seek to learn the dynamics of the SWE and predict the water height $h(x,t)$ and the momentum $uh(x,t)$. Therefore, we generate several testing datasets with various levels of the total water height and averaged velocity to assess how the dynamics of the SWE depends on these large-scale quantities.
In particular, we consider several different perturbation levels for $s_h$ and $s_u$ as summarized in Table \ref{test}. 
These datasets have different sets of initial conditions, but within each 
dataset the "true" and "predicted" trajectories are generated using the same set of $J=20$ different initial conditions.
$\text{TEST}_0$ is the reference testing dataset, since for the training data $h_0=4$
and $u_0=2.5$ ($s_h=s_u=0$).
The initial state of the reservoir is $r(0) = 0$ in all tests.

For each initial condition, $X^{(i)}(0)$ with $i=1,\ldots,J$, 
we generate one "true" trajectory, $X^{(i)}_{true}(t)$ and one "prediction" trajectory $X^{(i)}_{pred}(t)$. The "true" trajectory $X^{(i)}_{true}(t)$ is generated using a high-fidelity simulation of the SWE and the "prediction" trajectory is generated using the RC time-stepping. 
{\colb
We would like to point out that while the time-step in high-fidelity simulations of the SWE is restricted by the CFL condition $\delta t < O(10^{-3})$
(exact constant depends on the particular trajectory), 
the ESN prediction is generated with the same time-step as the training data, $\Delta t=0.1$.

{\bf Transfer Learning Data.} 
Data in the source domain $\calD_s$ consists of $M=20$ trajectories on $t\in[0,20]$ in the $\text{TEST}_0$ dataset computed from high-fidelity simulations of the SWE with initial conditions \eqref{IC_train} and sampled with $\Delta t=0.1$.
Transfer learning data in the target domain $\calD_t$
consists of one trajectory on $t \in [0,t^*]$ 
computed using high-fidelity simulations of the SWE 
and sampled with time-step $\Delta t=0.1$. 
This trajectory is computed 
using initial conditions \eqref{IC_test} 
consistent with each testing dataset $\text{TEST}_j$ in Table \ref{test}.
For majority of results presented here we use $t^*=10$, but we also compare how prediction error depends on $t^*$ for the testing data $\text{TEST}_8$ in Figure \ref{fig:test8_compare}.

We would like to point out that when $\alpha=0$ there is no knowledge transfer from the source domain, but
the target training data is insufficient to produce an adequate prediction. In addition, when $\alpha \to \infty$ ($\delta W=0$ since there is a strong penalty on $\delta W$), prediction errors are much larger than with transfer learning (depicted in Figures below).

Transfer learning can be sensitive to the particular value of the learning parameter $\alpha$ in \eqref{eq:deltaw}. Transfer learning parameter $\alpha$ should balance the norms in \eqref{eq:deltaw1}
so that the norm of the correction is approximately in the range 
$||\delta W||_\infty \approx [0.05,\ldots,0.2] \times ||W_{out}||_\infty$.
We calculated empirically that the optimal value of parameter 
$\alpha \in [10^{-7},\ldots,10^{-6}]$ for the amount of transfer learning data used here. We use $\alpha=5 \times 10^{-7}$ in all examples below.
We also would like to point out that $\alpha$
should change linearly with the amount of data in the target domain. 
This approach preserves the balance between the two norms in \eqref{eq:deltaw1}.
The connectivity and input matrices are identical for all ESNs with or without trnasfer learning. Please note that there is no transfer learning for the $\text{TEST}_0$ reference testing dataset. 
}
\begin{table}[h!]
\begin{center}
\begin{tabular}{|c|c|c|c|c|c|}
\hline
    & $J$ & $h_0+s_h$ &  $u_0+s_u$ & $L^2$ err $h+z$ NO TL & $L^2$ err $h+z$ WITH TL  \\
\hline
$\text{TEST}_0$ & 20 & 4.0 & 2.5  &  &  \\    
\hline
$\text{TEST}_1$ & 20 & 4.0 &  2.375  & 0.0034 & 0.0012 \\    
\hline
$\text{TEST}_2$ & 20 & 4.0 &  2.625  & 0.0025 & 0.0013 \\   
\hline
$\text{TEST}_3$ & 20 & 4.0 &  2.25  & 0.0076 & 0.002 \\    
\hline
$\text{TEST}_4$ & 20 & 4.0 &  2.75  & 0.0059 & 0.0014 \\    
\hline
$\text{TEST}_5$ & 20 & 3.9 &  2.5  & 0.0299 & 0.0011 \\   
\hline
$\text{TEST}_6$ & 20 & 4.1 &  2.5  & 0.0268 & 0.0013 \\    
\hline
$\text{TEST}_7$ & 20 & 3.8 &  2.5  & 0.0585 & 0.0012 \\    
\hline
$\text{TEST}_8$ & 20 & 4.2 &  2.5  & 0.0491 & 0.0012 \\   
\hline
$\text{TEST}_9$ & 20 & 4.4 &  2.5  & 0.076 & 0.0013 \\
\hline
\end{tabular}
\caption{Parameters of testing datasets for ESN with different initial mean water height $h_0 + s_h$ and initial mean velocity $u_0+s_u$.
Testing dataset $\text{TEST}_0$ corresponds to the ambient large-scale quantities for the training dataset ($h_0=4$, $u_0=2.5$). Right two columns represent averaged $L^2$ errors for the prediction of the total water height $h+z$ without and with transfer learning. Errors are averaged over $J=20$ trajectories and over the time interval $[0,20]$.}
\label{test}
\end{center}
\end{table}

\textbf{Prediction Error.}
{\colb
Since our goal here is to assess how well the ESN is able to learn the dynamics of the shallow-water equations, we measure the performance of the ESN by computing $J$ testing trajectories and  
use the $L^2$ error averaged over the whole set of testing trajectories as a measure of prediction utility. We compute averaged 
$L^2$ errors for each testing dataset in Table \ref{test} separately. In addition, we also plot selected solution snapshots.} 
Thus, for the set of trajectories in each $\text{TEST}_j$ we compute $L^2$ errors between the "true" and "prediction" trajectory with the same initial conditions 
\[
    e_{L^2}^{(i)}(t) = \frac{||X^{(i)}_{true}(t) - X^{(i)}_{pred}(t)||_{L^2}}{<||X^{(i)}_{true}(t)||_{L^2}>},
\]
where $i=1,\ldots,J$ denotes the trajectory index.
Here $< \cdot>$ denotes averaging over the time interval for prediction. 
We then define the prediction error as
\begin{equation}
\label{l2error}
    E_{L^2}(t)= \frac{1}{J} \sum_{i=1}^{J} {e_{L^2}^{(i)}(t)},
\end{equation}
where $J$ is the number of trajectories in a particular testing dataset. 
We use the average prediction error in \eqref{l2error} as a main criteria to assess how well the dynamics of the SWE is represented by the ESN. Note, that $E_{L^2}(t)$ also implicitly depends on parameter $\alpha$ for testing datasets $\text{TEST}_j$ with $j=1,\ldots,9$.

\section{Results}
\label{sec:results}

\begin{figure}[h]
    \centering
    \centerline{\includegraphics[width=0.5\textwidth, height=0.3\textwidth]{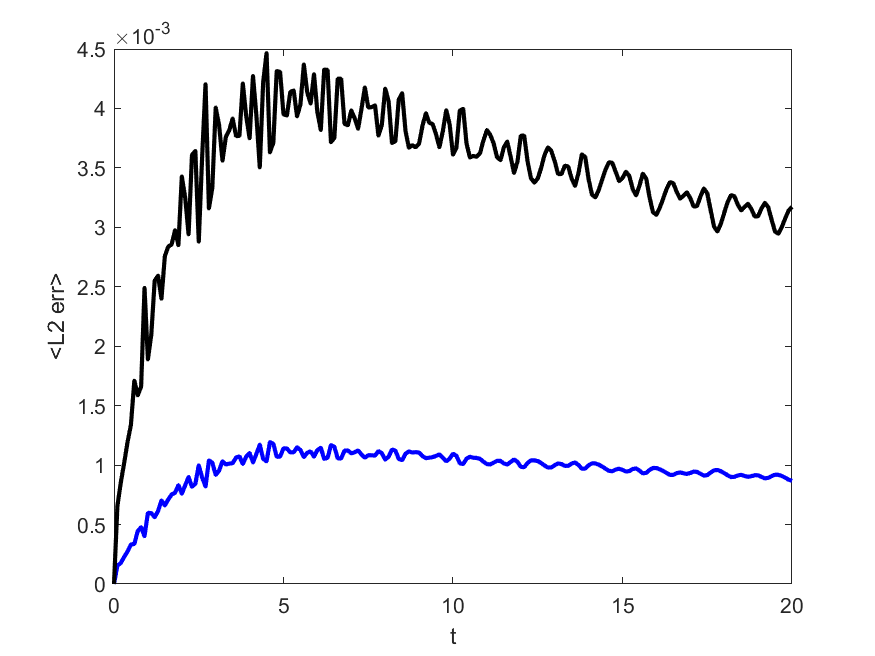}
    \includegraphics[width=0.5\textwidth, height=0.3\textwidth]{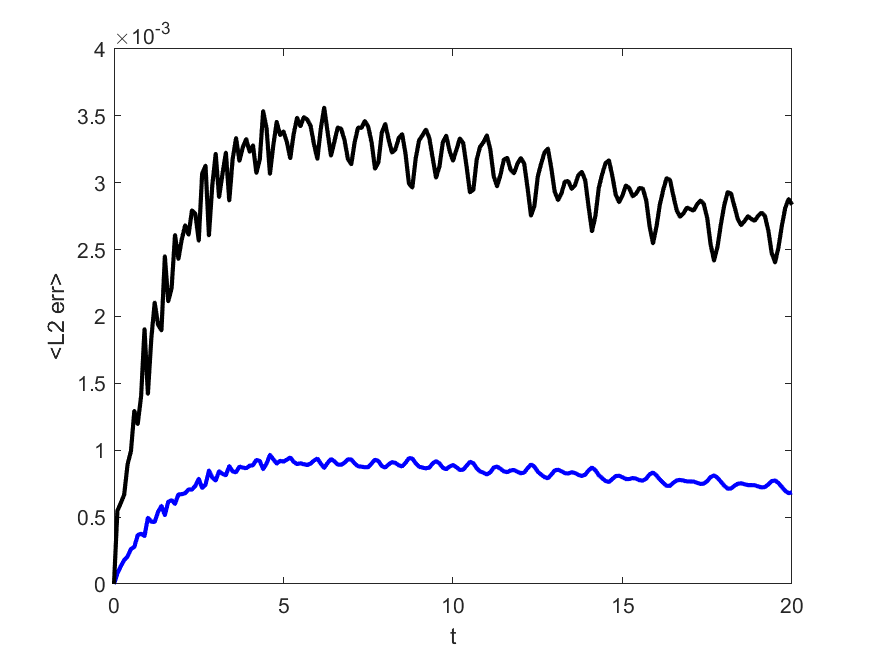}}
    \centerline{\includegraphics[width=0.5\textwidth, height=0.3\textwidth]{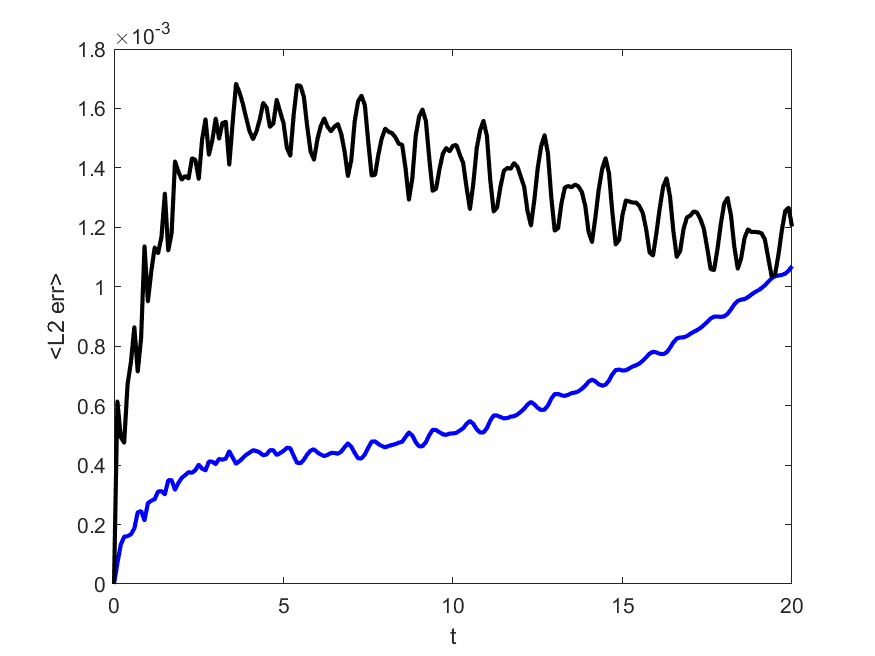}
    \includegraphics[width=0.5\textwidth, height=0.3\textwidth]{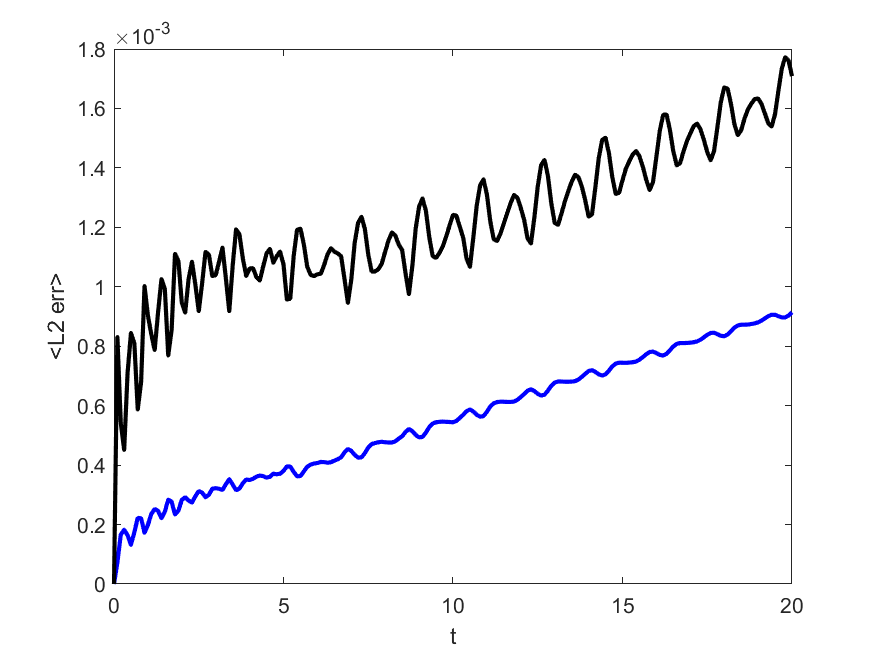}}
    \caption{Averaged $E_{L^2}(t)$ error in ESN predictions for the total water height $h + z$ (blue line)
    and momentum $hu$ (black line) of $\text{TEST}_0$ with different ESNs.
    Here $W_{in}$ are generated using four different values of $\beta_1$.
    Top left - $\beta_1=0.01$, top right - $\beta_1=0.04$, bottom left - $\beta_1=0.1$, bottom right - $\beta_1=0.2$.}
\label{fig:test0}
\end{figure}

We can expect that ESN has the best performance for predicting trajectories for dataset $\text{TEST}_0$ since this dataset contains trajectories with the same ambient large-scale quantities as the training data.
Averaged $L^2$ errors for predictions of the total water height $h+z$ are depicted in Figure \ref{fig:test0}. We obtain similar results for the momentum.
{\colb
We can see that prediction errors are quite small (less than one percent) on the whole interval $[0,20]$. We can see that prediction errors depend weakly on the parameter $\beta_1$ which is used in the generation of $W_{in}$. Thus, we use $\beta_1=0.1$ for the rest of the paper, but prediction results are quantitatively similar for a range of $\beta_1 \in [0.01,\ldots,0.2]$.
We also obtained comparable results with varying $\beta_1$ for 
prediction trajectories for the $\text{TEST}_8$ dataset with transfer learning.
}

\begin{figure}[h!]
    \centering
    \includegraphics[width=\textwidth]{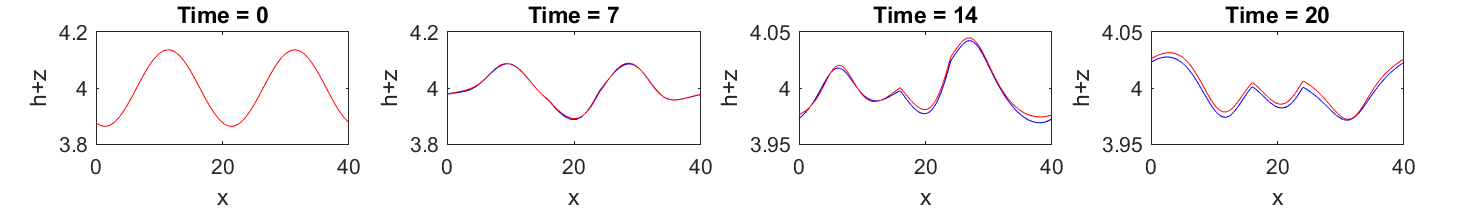}
    \includegraphics[width=\textwidth]{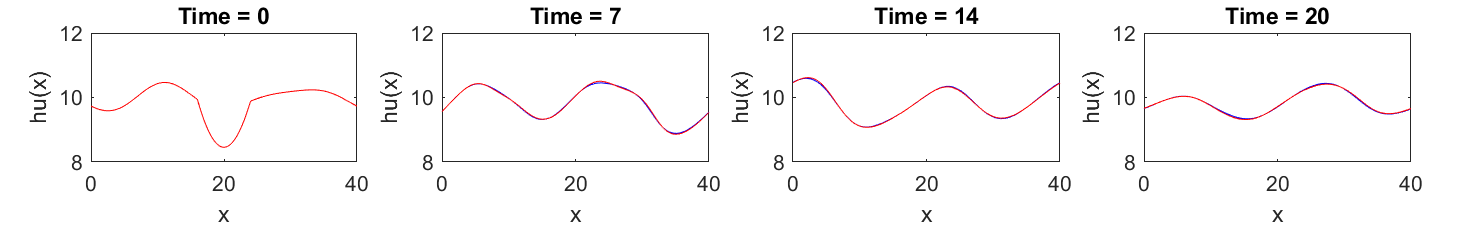}
    \caption{Comparison of individual snapshots for the total water height (top) and momentum (bottom)
    for one particular trajectory from the $\text{TEST}_0$ testing dataset.
    Red lines - DNS of the SWE, 
    Blue lines - predictions of the ESN.}
    \label{fig:test0_snap}
\end{figure}

\begin{figure}[h]
\centerline{
            \includegraphics[width=0.5\textwidth, height=0.3\textwidth]{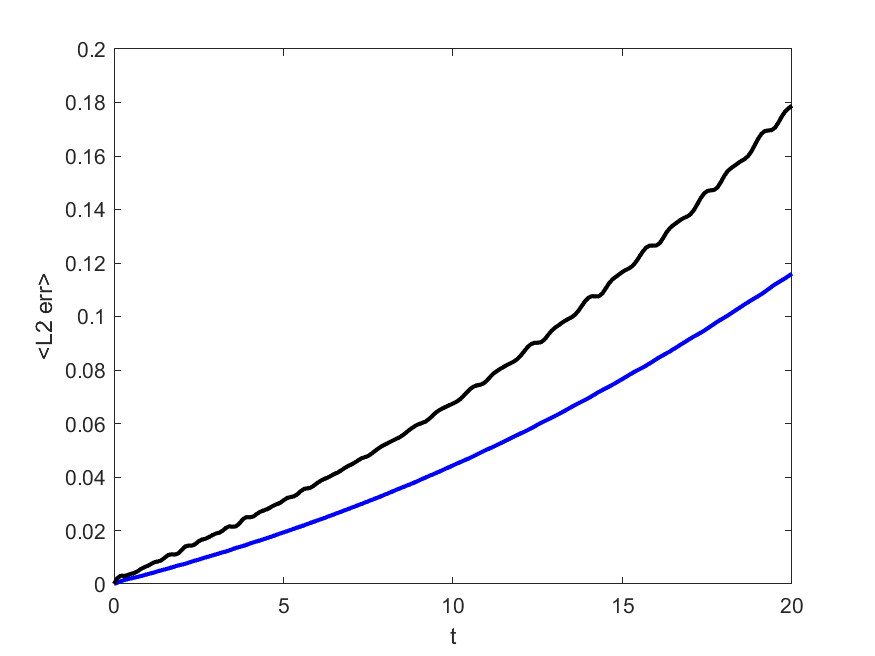}
            \includegraphics[width=0.5\textwidth, height=0.3\textwidth]{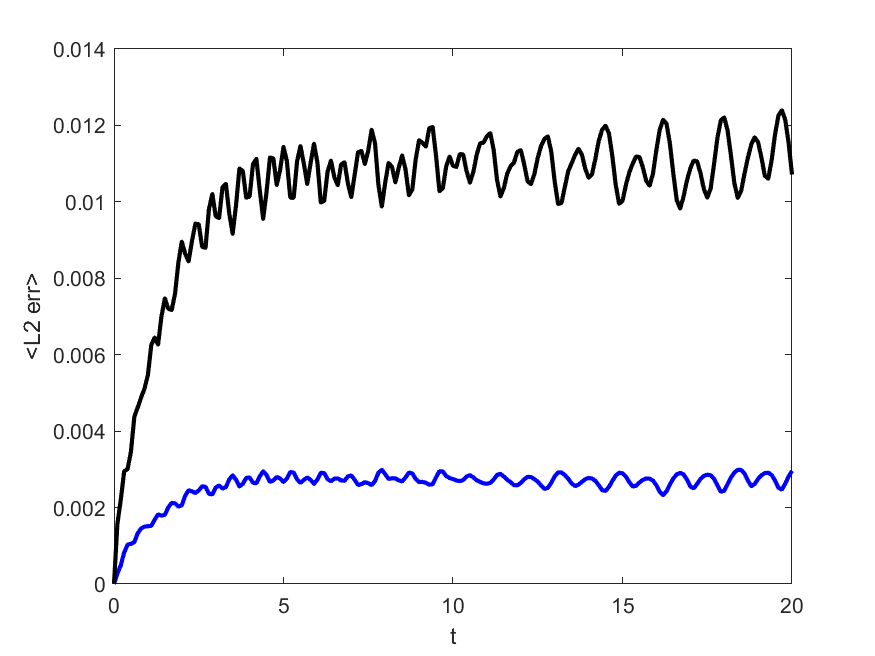}
}
\centerline{
            \includegraphics[width=0.5\textwidth, height=0.3\textwidth]{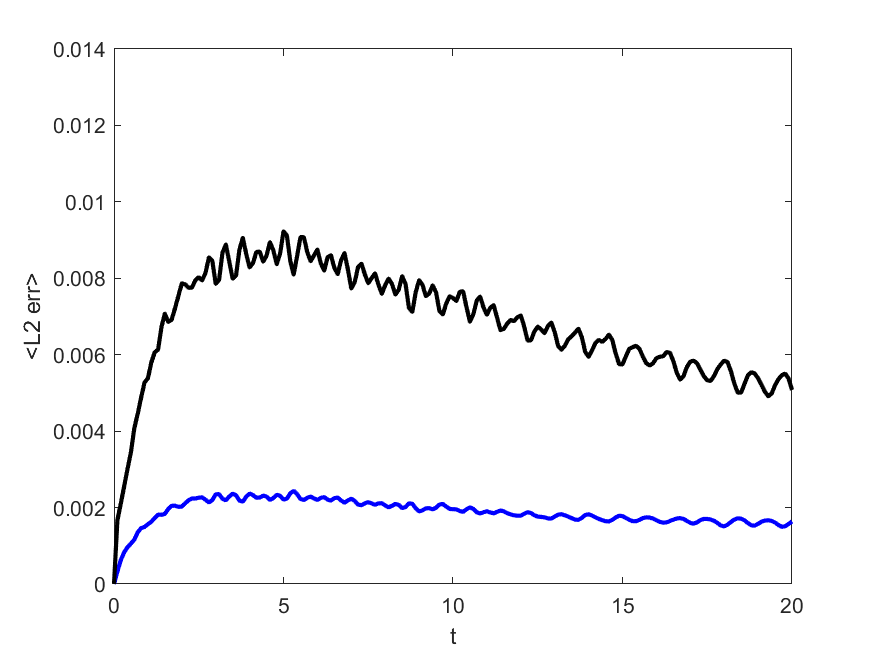}
            \includegraphics[width=0.5\textwidth, height=0.3\textwidth]{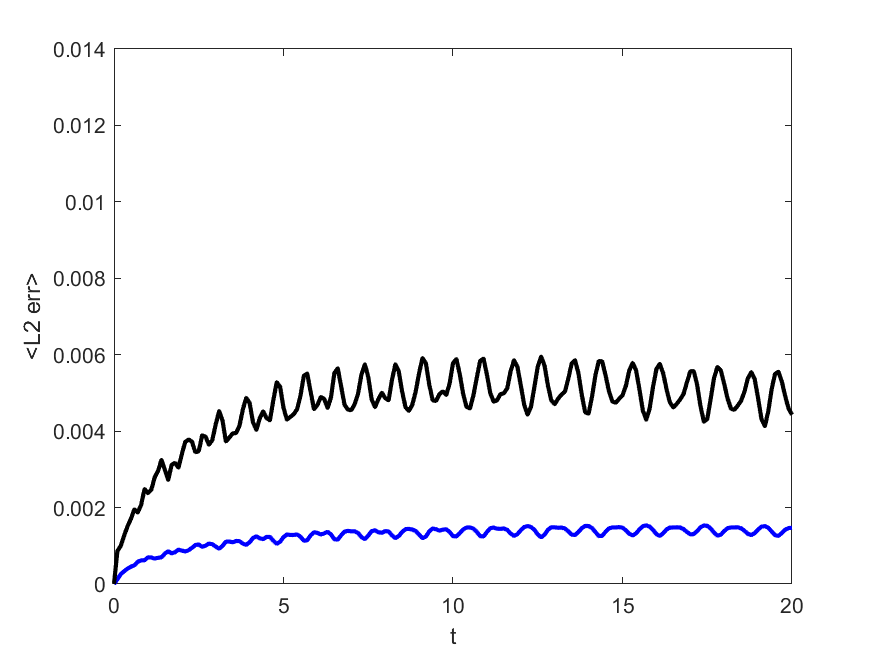}
}
    \caption{Averaged ${L^2}(t)$ errors in ESN predictions of the total water height $h+z$ (blue line) and momentum $hu$ (black line) 
    for $\text{TEST}_8$ dataset with four different ESNs. 
    Upper left - no transfer learning, upper right - transfer learning with data on $t\in[0,2]$, 
    lower left - transfer learning with data on $t\in[0,5]$,
    lower right - transfer learning with data on $t\in[0,10]$.
    Errors are averaged over $J=20$ trajectories. Notice the difference between the vertical scale for the plot without transfer learning, and plots with transfer learning.}
\label{fig:test8_compare}
\end{figure}

High accuracy of ESN predictions is confirmed by individual snapshots depicted in Figure \ref{fig:test0_snap}.
Overall, ESN has a high utility of prediction in the regime it was originally trained.
However, performance of the ESN deteriorates for initial conditions
which are not consistent with the training regime. We observe that prediction errors increase considerably for testing datasets 1-9.
Moreover, performance of the ESN affected stronger for initial conditions with shifts in the water level (i.e. $s_h \ne 0$).
Thus, for each testing dataset we employ transfer learning and generate a small dataset $\bX^*$ from the target domain by performing direct numerical simulations \emph{for a single trajectory} on $t\in[0,10]$ of the shallow water equations with an initial condition in \eqref{IC_test} consistent with the ambient large-scale quantities in the testing data (i.e. corresponding levels of $s_h$ and $s_u$). 
We sample this trajectory with time-step $\Delta t=0.1$. 
The learning rate is chosen as $\alpha = 5 \times 10^{-7}$.
We depict averaged (over all trajectories in the dataset) prediction errors 
for $\text{TEST}_8$ dataset in Figure \ref{fig:test8_compare}. 
ESN prediction without the transfer learning ($\alpha=\infty$) leads to very large errors over a very short time. Therefore, it is crucial to use transfer learning to take into account large-scale quantities such as the total water height and velocity.
We also observe that the dataset $\mathcal{\bX^*}$ is insufficient to train the ESN from scratch in the new regime and obtain adequate prediction results without transfer learning ($\alpha=0$). 
{\colb
In addition, Figure \ref{fig:test8_compare} also depicts how prediction errors depend on the size of the transfer learning dataset $\bX^*$. In particular, we consider four cases - no transfer learning and transfer learning with three datasets sampled 
over $t\in[0,t^*]$ with $t^*=2,5,10$.
Prediction errors decay significantly 
with transfer learning, and even a small target domain dataset
sampled on $t\in[0,2]$ significantly improves ESN predictions. Large errors for prediction without transfer learning can be primarily attributed to a vertical systematic shift of the solution. However, we also comment later in this section that for some initial conditions simple vertical adjustment of the ESN prediction without transfer learning would not provide an accurate estimate of the high-resolution simulations of the SWE.  
}

\begin{figure}[h]
\centerline{
            \includegraphics[width=0.5\textwidth, height=0.3\textwidth]{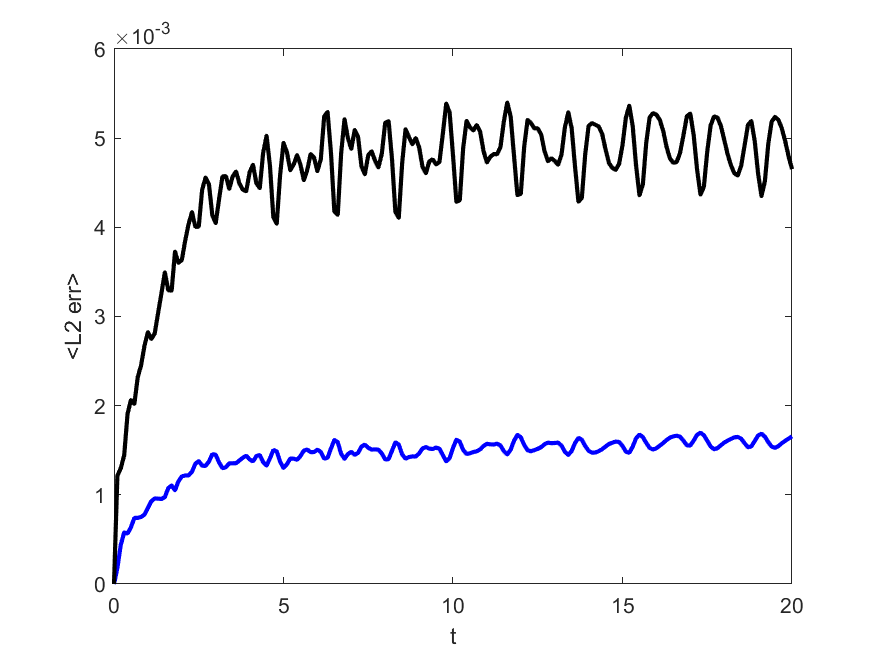}
            \includegraphics[width=0.5\textwidth, height=0.3\textwidth]{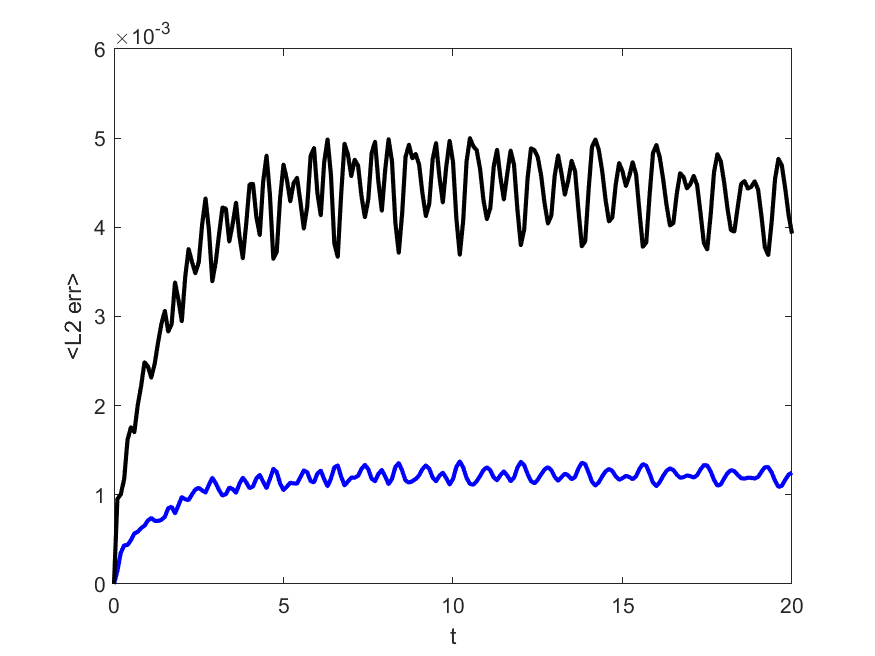}
}
\centerline{
            \includegraphics[width=0.5\textwidth, height=0.3\textwidth]{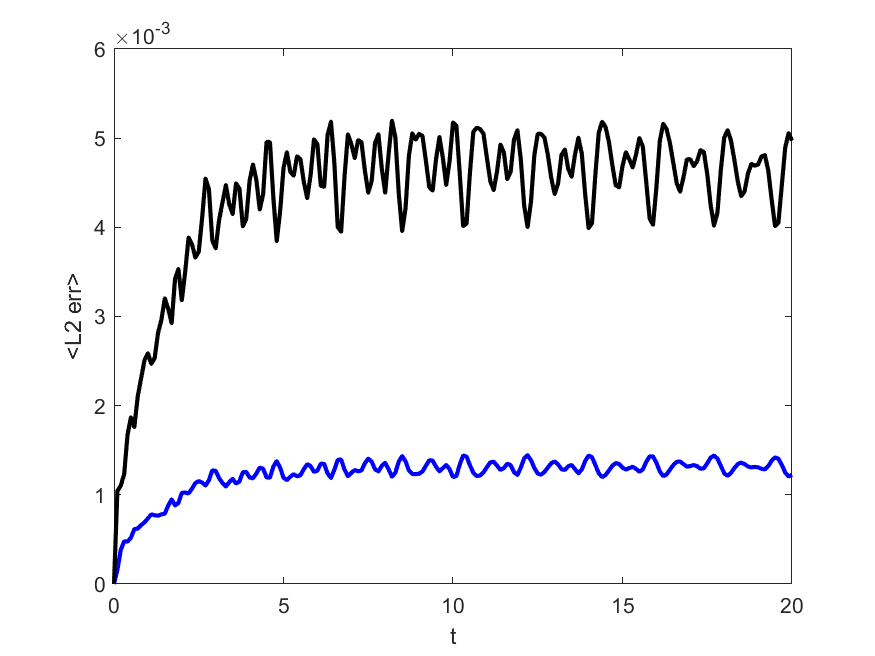}
            \includegraphics[width=0.5\textwidth, height=0.3\textwidth]{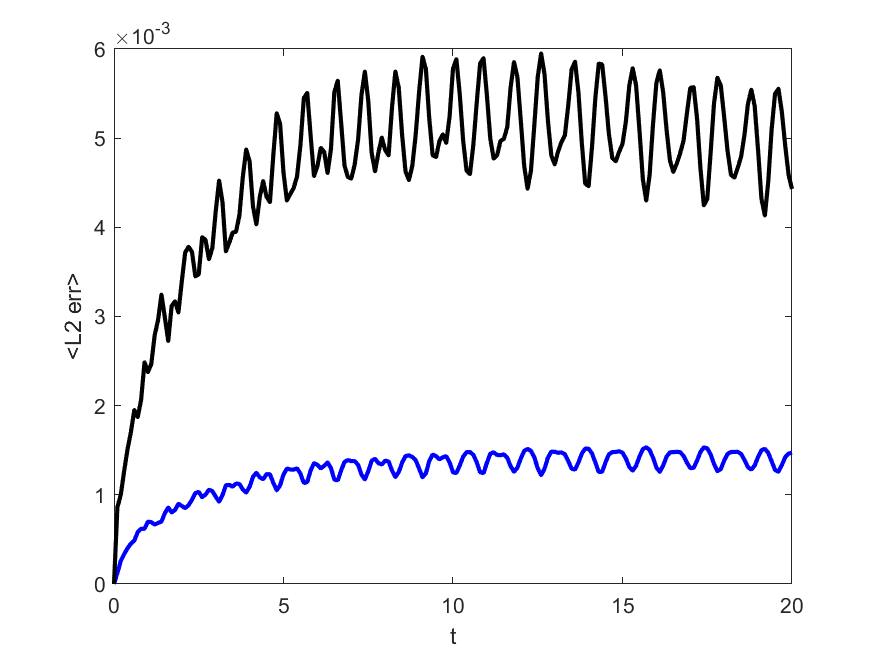}
}
    \caption{Averaged ${L^2}(t)$ errors in ESN predictions of the total water height $h+z$ (blue line) and momentum $hu$ (black line) with transfer learning 
    for $\text{TEST}_4$ (upper left),
    $\text{TEST}_5$ (upper right), $\text{TEST}_7$ (lower left), $\text{TEST}_8$ (lower right) testing datasets. Errors are averaged over $J=20$ trajectories.}
\label{fig:test3478_compare}
\end{figure}

\begin{figure}[h!]
    \centering
    \includegraphics[width=\textwidth]{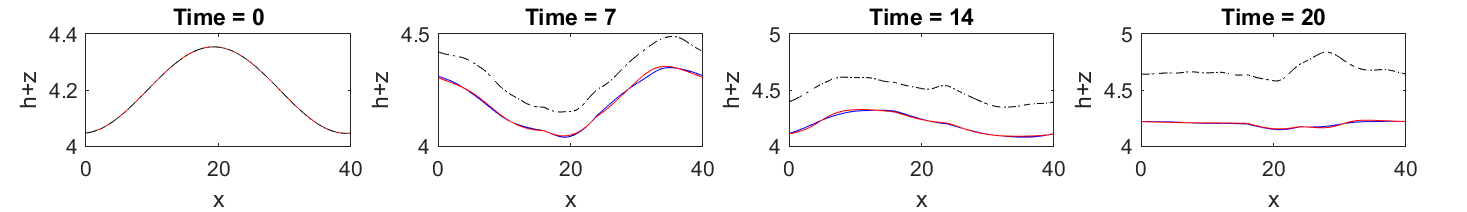}
    \includegraphics[width=\textwidth]{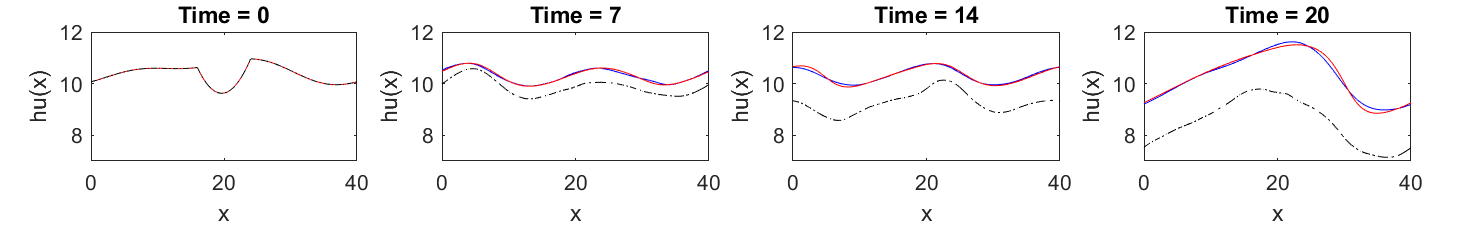}
    \caption{Comparison of individual snapshots for the total water height (top) and momentum (bottom)
    for one particular trajectory from the $\text{TEST}_8$ testing dataset.
    Red lines - DNS of the SWE, 
    Blue lines - predictions of the ESN with transfer learning, 
    Black dash-dot lines - predictions of the ESN without transfer learning.}
    \label{fig:test8snap}
\end{figure}

\begin{figure}[h!]
    \centering
    \includegraphics[width=\textwidth]{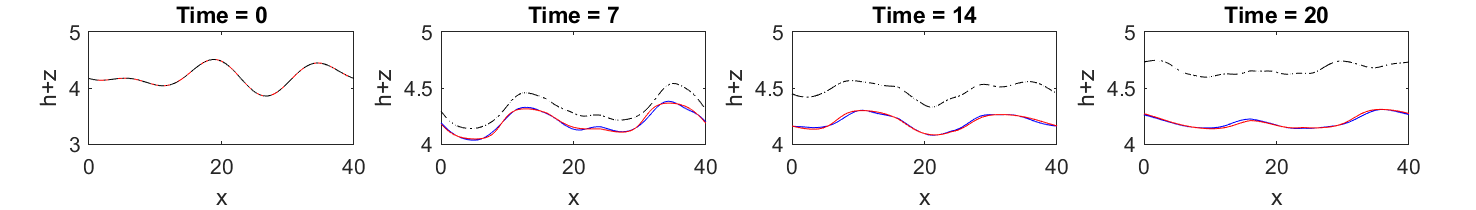}
    \includegraphics[width=\textwidth]{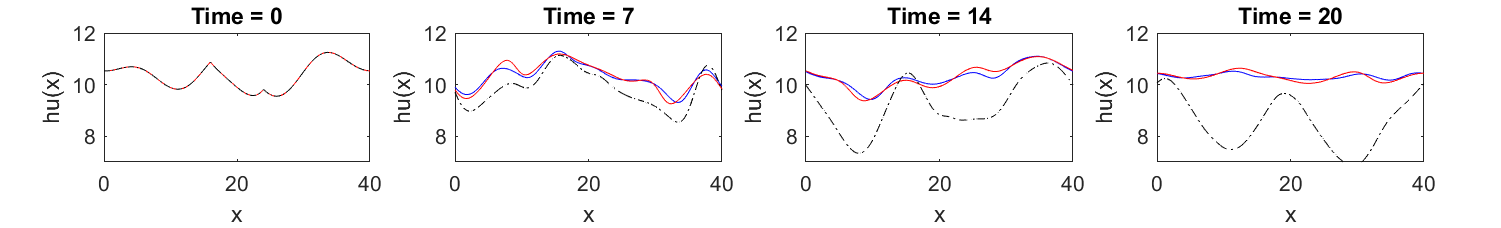}
    \caption{Comparison of individual snapshots for the total water height (top) and momentum (bottom)
    for one particular trajectory from a testing dataset with $s_u=0$ and $s_h=.2$, 
    but with two perturbation frequencies for the initial conditions of 
    both the total water height and the velocity. The initial conditions are similar to 
    \eqref{IC_test}, except there is a second trigonometric term in the initial condition for both, the total water height and the velocity.
    In particular, perturbation frequencies are $k_{1,2}=3,2$ and $p_{1,2} = 3,1$ 
    for the trajectory depicted here.
    Red lines - DNS of the SWE, 
    Blue lines - predictions of the ESN with transfer learning, 
    Black dash-dot lines - predictions of the ESN without transfer learning.}
    \label{fig:test82fsnap}
\end{figure}

\begin{figure}[h!]
    \centering
    \includegraphics[width=\textwidth]{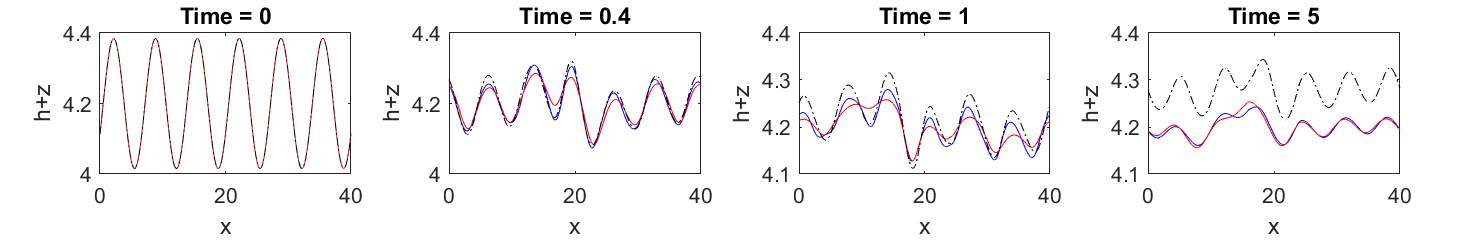}
    \caption{Comparison of individual snapshots for the total water height (top) and momentum (bottom)
    for one particular trajectory from a testing dataset with $s_u=0$ and $s_h=.2$, 
    but with higher perturbation frequencies for the initial conditions of 
    both the total water height and the velocity. 
    In particular, $k=6$, $p=2$ in the initial condition \eqref{IC_test} for the trajectory depicted here.
    Red lines - DNS of the SWE, 
    Blue lines - predictions of the ESN with transfer learning, 
    Black dash-dot lines - predictions of the ESN without transfer learning.}
    \label{fig:test8hfsnap}
\end{figure}

{\colb
Transfer learning leads to a considerable improvement of ESN predictions.
In particular, prediction errors for ESN models with transfer learning are under 5\% for all simulated regimes.
This is slightly less than one order of magnitude improvement compared to 
predictions of the ESN without transfer learning.
Averaged prediction errors for 4 selected datasets are presented in Figure \ref{fig:test3478_compare} 
and averaged (averaged over $J=20$ trajectories and over time) 
prediction errors for all testing datasets with and without transfer learning are presented in Table \ref{test}. 
Snapshots of selected individual 
trajectory from $\text{TEST}_8$ dataset are presented in Figure 
\ref{fig:test8snap}. Prediction of ESN without transfer learning significantly overestimates the total water level. While for times $t=7,14$ the correct prediction for the water level (red line) can be obtained by shifting the trajectory without transfer learning down, for time $t=20$ predicted trajectory of the ESN without transfer learning exhibits a nonlinear feature which is absent in the high-resolution simulation of the SWE.

Overall, ESN prediction without transfer learning can often be corrected by "shifting" the predicted total water height to an appropriate level. However, 
there are approximately 20-25\% of initial conditions where this simple adjustment would not produce accurate predictions by the ESN without transfer learning. These initial conditions often correspond to large initial perturbation (e.g. $a\approx 0.05$)
and high frequency in the Fourier space (e.g. $k\ge 4$).
Moreover, 
we would like to point out that nonlinear features of the ESN are often manifested stronger for the momentum (or velocity). In particular, we performed additional tests where initial conditions were chosen similar to \eqref{eq:predict} with $s_u=0$ and $s_h=.2$ (same as for the $\text{TEST}_8$ dataset), but with two perturbation frequencies for both the total water height and the velocity, i.e. $k_1, k_2, p_1, p_2 \sim Uniform(1,2,3,4)$ with additional trigonometric terms for frequencies $k_2$ and $p_2$. Snapshots of one particular trajectory are presented in Figure \ref{fig:test82fsnap}.
These results demonstrate that the momentum snapshots for times $t=14,20$ are completely misrepresented by the ESN without transfer learning.

For initial conditions with higher perturbation frequencies $k, p > 4$, the SWE dynamics is "smoothed-out" faster than the prediction of the ESN. Thus, for short times  prediction of the ESN appears more oscillatory than the dynamics of the SWE. In particular, prediction of the ESN with transfer learning for initial conditions where the frequency of perturbation for $h(x,0)+z(x)$ is $k=6$ is depicted in Figure \ref{fig:test8hfsnap}. Here we also observe that the numerical solution of the SWE cannot be accurately predicted by ESN without transfer learning even if the ESN prediction is shifted to an appropriate level.
}

Our results demonstrate that it is more challenging to predict 
trajectories of the SWE with initial conditions where the mean water height and/or mean velocity are different from the training regime. Transfer learning is able to alleviate these problems and accuracy of predictions with transfer learning is approximately equal for initial conditions with a different mean water height 
($\text{TEST}_5$ - $\text{TEST}_8$ datasets in Table \ref{test} and bottom row Figure \ref{fig:test3478_compare})
and mean velocity
($\text{TEST}_1$ - $\text{TEST}_4$ datasets in Table \ref{test} and top row Figure \ref{fig:test3478_compare}).


\section{Conclusions}
In this paper we present a modification of the ESN machine learning approach. 
ESNs have been successful in predicting future evolution of trajectories of chaotic and turbulent systems in stationary regimes after they have been trained on earlier evolution of the same trajectory.
However, in this work we place particular emphasis on learning the dynamics of the underlying equations, i.e. we develop an approach capable of predicting transient behavior of trajectories of the shallow-water equations with arbitrary initial conditions.
In this sense our method adds to the growing number of ML methods that have focussed on data-driven discovery of partial differential equations (e.g., \cite{rudy2017data, long2018pde, berg2019data, reinbold2019data, champion2019data, degennaro2019model, xu2019dl, maslyaev2020data, li2020robust, zhang2021robust} and others).

We demonstrate that the ESN is capable of learning the dynamics of the SWE and generates high-utility predictions for initial conditions which are consistent with the ambient large-scale quantities in the training data. An important practical aspect is that the training dataset can include concatenated trajectories. Thus, the training data can be generated in parallel, which significantly accelerates data-generation for training. In addition, we also demonstrated that ESN is able to generate high-utility predictions with the time-step which is much larger than the time-step in direct numerical simulations (i.e. $\Delta t \gg \delta t$), which also has practical significance. 

{\colb
ESN training only involves computing the output matrix $W_{out}$. With the reservoir size $D=4800$ and training dataset $\bX$ of the size $800 \times 4000$ (20 trajectories sampled on $t\in[0,20]$ with $\Delta t=0.1$), training takes approximately 90 seconds on an Intel i7-7500 2.70GHz laptop with 16GB memory. Calculations are done with the standard numpy library. The same laptop is able to handle reservoirs up to size $D=10,000$, but larger reservoir size does not lead to an improved performance by the ESN. Testing of 20 trajectories with transfer learning takes approximately the same time.
}

{\colb Reservoir computing has been applied to much more complex fluid dynamical models,
such as a realistic low-resolution weather prediction model \cite{arcomano2020machine} and the 
2D Boussinesq equations \cite{pasch20}. Therefore, we expect that reservoir computing can be applied to a 2D shallow-water model as well.
However, it might be beneficial to combine the reservoir computing ML model with 
Principal Component Analysis (PCA) to extract the leading Empirical Orthogonal Functions in a suitable basis and, thus, reduce the dimensionality of the data (see e.g. \cite{pasch20}).
Time-dependent coefficients of principal components can then be arranged as a one-dimensional array, and the same reservoir computing formalism (as discussed in this paper) can be applied to these coefficients.
An alternative Machine Learning approach
would be to apply the encoder/decoder combination to the solution of the 2D SWE to reduce the dimensionality of the data.
}

Performance of the ESN deteriorates for initial conditions with "shift" in the averaged water height or velocity. Therefore, it is essential to utilize transfer learning to update the ESN model for a parameter regime which is not consistent with the training data. Here we were able to successfully 
use a small dataset (single trajectory on $t \in [0,10]$) to update the output matrix $W_{out}$ and significantly improve predictions generated by the ESN. Even transfer learning on $t \in [0,2]$ leads to significant improvement.
{\colb
The transfer learning approach discussed here works well for "shifts" in the initial averaged water height within $\pm$10\%. The transfer learning algorithm requires larger transfer learning datasets $\bX^*$
for larger "shifts" in the water height. However, such datasets can consist of multiple trajectories with random initial conditions (generated with the same "shifted" averaged water height $h_0$) and, thus, can be generated in parallel. We also would like to comment that the performance of the transfer learning is less sensitive to "shifts" in the averaged initial velocity.}
In general, predictions can be improved by using larger transfer learning datasets with several concatenated trajectories. Since such trajectories can be generated in parallel, this does not significantly increase the data-generating time for the transfer learning step.
Our approach significantly accelerates generating ``prediction'' trajectories. The resulting algorithm (including generating data for transfer learning and updating the output matrix) is approximately 5-8 times faster than direct numerical simulations. This is due to a larger sampling and prediction time-step, $\Delta t$.
Results presented here imply that the dynamics of the shallow water equations depends on the ambient large-scale parameters in a continuous fashion. Moreover, small changes in the ambient parameters 
probably correspond to relatively small changes in trajectories and, therefore, small changes in the output matrix $W_{out}$.

The machine learning approach developed here can also be potentially utilized to track changes in the behavior of complex turbulent systems with respect to equations' parameters. In particular, it should be sufficient to generate a small new dataset to update the output matrix $W_{out}$ and significantly improve the predictive power of the reservoir computing model in a new parameter regime. A sufficient condition for the success of the transfer learning approach in this context
is related to classical results for continuous dependence of solutions on parameters. However, the transfer learning approach may not be suitable for predicting trajectories of a system near a bifurcation points since the model's behavior typically changes drastically during a bifurcation. This will be investigated in a subsequent paper.

Overall, the reservoir computing approach with transfer learning is an efficient approach for generating prediction trajectories in a non-equilibrium setting. This approach has many potential applications, 
including numerical prediction techniques utilizing ensemble simulations.
{\colb 
We demonstrate that RC is highly efficient computationally since it is not restricted by the CFL condition 
and is able to generate trajectories with a much larger time-step compared with numerical simulations of 
the SWE. Moreover, the ESN model can be used in parallel to quickly generate much larger ensembles of trajectories compared to direct numerical simulations. 
We expect that the acceleration factor will be even larger for higher-dimensional problems if ESN is combined with PCA analysis or machine learning techniques for solution compression 
(e.g. variational autoencoders).
}

\section*{Acknowledgements}
Research of I. Timofeyev and X. Chen has been partially supported by grants NSF DMS-1620278 and ONR N00014-17-1-2845.
Nadiga was supported under DOE's SciDAC program under SciDAC4 project ``Non-Hydrostatic Dynamics with Multi-Moment Characteristic Discontinuous Galerkin Methods (NH-MMCDG).''

{\colb
The model code and data for this project are publicly available online at github in the repository Timofeyev/SWE\_ESN and on zenodo under the title "Predicting Shallow Water Dynamics using Echo-State Networks with Transfer Learning" or
DOI: https://doi.org/10.5281/zenodo.6828772
}


\end{document}